\documentclass[10pt,twocolumn,letterpaper]{article}

\usepackage{cvpr}
\usepackage{times}
\usepackage{epsfig}
\usepackage{graphicx}
\usepackage{amsmath}
\usepackage{amssymb}
\usepackage{enumerate}
\usepackage{bm}
\usepackage{amsmath}
\usepackage{nccmath}

\usepackage[pagebackref=true,breaklinks=true,letterpaper=true,colorlinks,bookmarks=false]{hyperref}

\cvprfinalcopy 

\def\cvprPaperID{5258} 

\ifcvprfinal\fi
\begin{document}

\title{A Context-and-Spatial Aware Network for Multi-Person Pose Estimation}

\author{
Dongdong Yu$^{1}$, Kai Su$^{1,2}$, Xin Geng$^{2}$, Changhu Wang$^{*,1}$ \\
$^{1}$ByteDance AI Lab, Beijing, China \\
{\tt\small \{yudongdong,sukai,wangchanghu\}@bytedance.com} \\
$^{2}$School of Computer Science and Engineering, Southeast University, Nanjing, China \\
{\tt\small \{sukai,xgeng\}@seu.edu.cn} \\
}
\maketitle

\begin{abstract}
   Multi-person pose estimation is a fundamental yet challenging task in computer vision. Both rich context information and spatial information are required to precisely locate the keypoints for  all persons in an image. In this paper, a novel Context-and-Spatial Aware Network (CSANet), which integrates both a Context Aware Path and Spatial Aware Path, is proposed to obtain effective features involving both context information and spatial information. Specifically, we design a Context Aware Path with structure supervision strategy and spatial pyramid pooling strategy  to enhance the context information. Meanwhile, a Spatial Aware Path is proposed to preserve the spatial information, which also shortens the information propagation path from low-level features to high-level features. On top of these two paths, we employ a Heavy Head Path to further combine and enhance the features effectively.  Experimentally,  our proposed network outperforms state-of-the-art methods on the COCO keypoint benchmark, which verifies the effectiveness of our method and further corroborates the above proposition.
\end{abstract}

\section{Introduction}
Multi-person pose estimation aims at locating body keypoints (eyes, ears, nose, shoulders, elbows, wrists, hips, knees, ankles, etc.) for all persons from an image. It is  fundamental and important to a variety of computer vision applications, such as human action recognition~\cite{wang2013approach} and human re-identification~\cite{liu2018pose}. 

\begin{figure}[t]
\begin{center}
\includegraphics[width = 8cm]{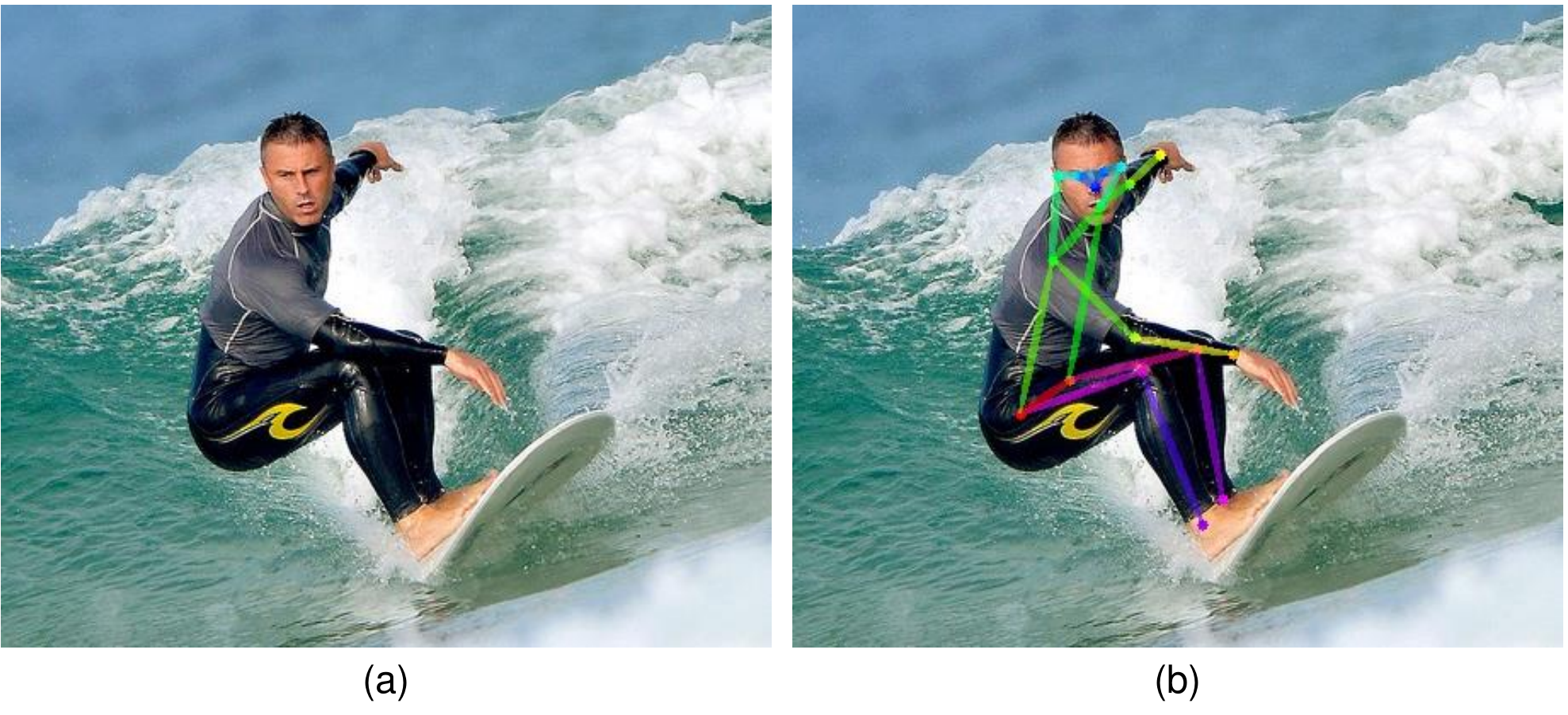}
\end{center}
   \caption{An occluded example from MS-COCO test-dev2017 dataset. (a) is the original input image. (b) is the prediction result of our CSANet.}
\label{fig:image1}
\end{figure}

Due to the help of deep convolution neural networks, remarkable progress has been made in multi-person pose estimation~\cite{insafutdinov2016deepercut,toshev2014deeppose,chen2017cascaded,xiao2018simple,newell2017associative,AAAI1817206,yang2017pyramid,papandreou2017towards,newell2016stacked,ke2018multi,cao2017realtime,chu2017multi}.  Although great progress has been made,  there still exist a lot of challenging cases, such as occluded  keypoints, invisible keypoints, change of view point, and crowed background. Both affluent context information and spatial information are essential to locate keypoints accurately. For example, we can capture the global context of the image by enlarging  the receptive field and fusing different context information. The context information represents the global position of human and indicates the contextual relationship between keypoints,  thus holds potential to accurately estimate the occluded and invisible keypoints, \textit{e.g.} the left knee of the man in Figure~\ref{fig:image1}. Adding the spatial information can provide detail information which is useful for refining the positions of keypoints. 

\begin{figure*}
\begin{center}
\includegraphics[height = 5cm]{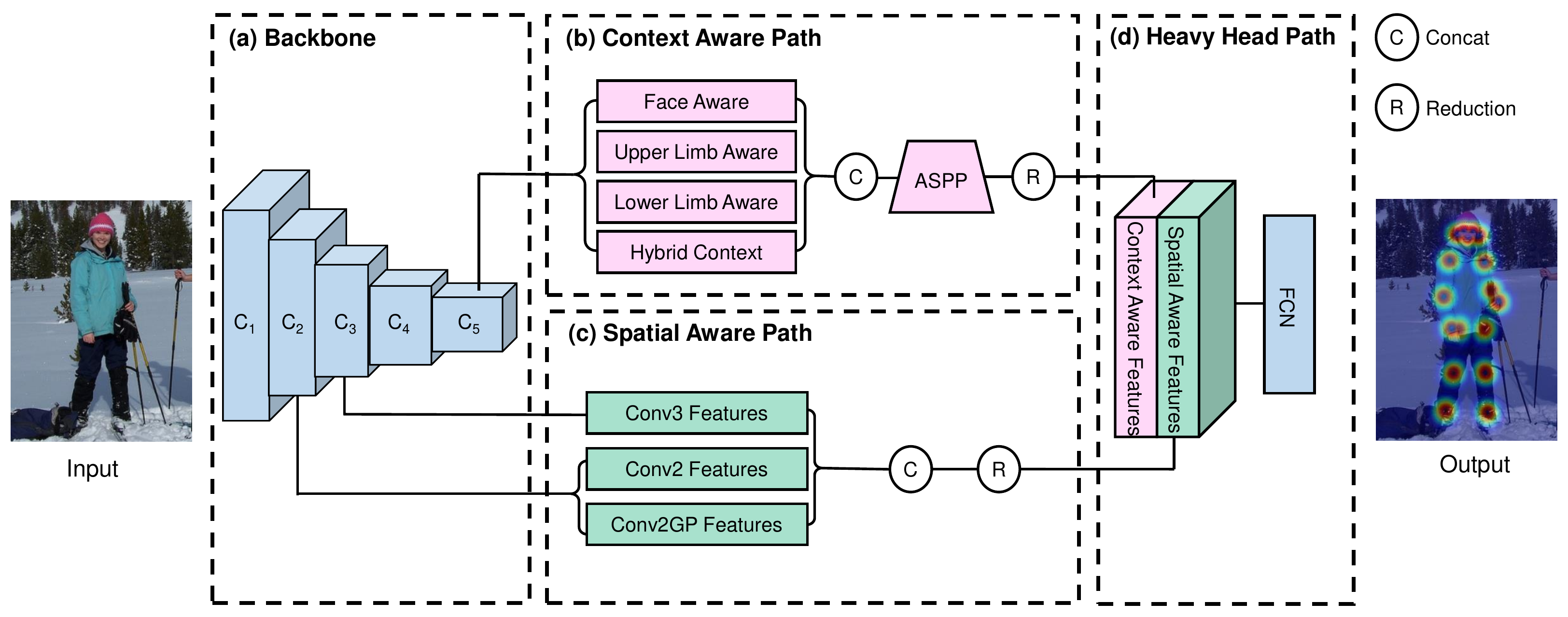}
\end{center}
   \caption{Illustration of our framework. (a) ResNet backbone. (b) Components of the Context Aware Path (CAP). (c)  Components of the Spatial Aware Path (SAP). (d) Components of the Heavy Head Path (HHP). Note that the reduction operation is implemented by a convolution layer of $1\times1$ kernel size.}
\label{fig:image2}
\end{figure*}

With these observations, we aim at leveraging both the context information and spatial information to improve multi-person pose estimation. Toward this end, we present a novel Context-and-Spatial Aware Network (CSANet) to  extract effective context information and spatial information, as shown in Figure~\ref{fig:image2}. Based on a backbone network, there are three parts in our network architecture: Context Aware Path (CAP), Spatial Aware Path (SAP), and Heavy Head Path (HHP). For Context Aware Path, we design a structure supervision module to learn the part-aware context information and adopt the Atrous Spatial Pyramid Pooling (ASPP) module~\cite{chen2018deeplab} to capture context information of different receptive fields. Thus, sufficient context information are extracted to infer the occluded and invisible keypoints. In respect of Spatial Aware Path, we preserve the spatial size and encode affluent spatial information for accurate localization, which also shorten the information path from low-level features to the high-level features. On top of these two paths, the Heavy Head Path (HHP) is proposed to adopt adaptivity fusion learning to combine the context information with spatial information and employ a small fully convolution network to recalibrate the fusion features.

Based on our Context-and-Spatial Aware Network (CSANet),  we address the multi-person pose estimation problem in a top-down pipeline. First, we apply the human detect network to obtain human detection bounding boxes. Then, the CSANet is adopted to locate body keypoints for each human bounding box. Next, ablation studies are conducted to demonstrate the effectiveness of the CAP path, SAP path and HHP path. Finally, we evaluate our proposed network on the COCO keypoint benchmark~\cite{cocodataset.org}, and the experimental results show that our proposed CSANet outperforms existing state-of-the-art methods. 

In summary, there are five contributions in our paper:
\begin{itemize}
\setlength{\itemsep}{0pt}
\setlength{\parsep}{0pt}
\setlength{\parskip}{0pt}
\item We design a Context Aware Path to learn the part-aware context information and context information of different receptive fields for inferencing the challenge keypoints.  
\item We design a Spatial Aware Path to preserve spatial detail information for refining the position of keypoints.
\item We design a Heavy Head Path to adaptively fuse the context information and the spatial information.  
\item Based on the Context Aware Path, Spatial Aware Path and Heavy Head Path, we propose a Context-and-Spatial Aware Network which can make full use of the context information and spatial information.
\item We evaluate our method on the COCO keypoint benchmark, and achieve state-of-the-art performance in multi-person pose estimation.
\end{itemize}

\section{Related Work}

Recently, lots of approaches based on Convolution Neural Network (CNN) have achieved high performance on different benchmarks of multi-person pose estimation~\cite{insafutdinov2016deepercut,toshev2014deeppose,chen2017cascaded,xiao2018simple,newell2017associative,AAAI1817206,yang2017pyramid,papandreou2017towards,newell2016stacked,ke2018multi,cao2017realtime}. Several principles  proposed for designing networks in scene parsing are also effective for our work, in which we pay more attention to the issue of context information extraction and spatial information preservation~\cite{chen2017rethinking,zhao2017pyramid,chen2018deeplab,wang2018understanding,yu2018bisenet}.

\noindent{\textbf{Multi-person Pose Estimation}}~~Recently, significant progress has been made in multi-person pose estimation with the development of CNN. In~\cite{cao2017realtime}, a real-time Convolution Pose Machine (CPM) is proposed to locate the body keypoints, and assemble the keypoints to individuals in the image with the learning part affinity fields (PAFs). Based on the ResNet backbone, the Simple Baseline Network (SBN)~\cite{xiao2018simple} employs a deconvolution head network to predict human keypoints. The spatial detail information is inevitably lost along the information propogation in CPM and SBN, which is useful for refining keypoints' localization. To avoid this problem, Newell \textit{et al.}~\cite{newell2017associative} integrate associate embedding with a stack-hourglass network to produce joint score heatmaps and  embedded tags for grouping joints into individual people. The Cascaded Pyramid Network (CPN)~\cite{chen2017cascaded} adopts the GlobalNet to learn a good feature representation and the RefineNet to further recalibrate the feature representation for accurate keypoint  localization. The Hourglass network and Cascade Pyramid Network preserve spatial features at each resolution by adding skip layers and capture sufficient context information for accurately inferencing both simple keypoints and challenge keypoints.

As mentioned in~\cite{chen2017cascaded,AAAI1817206}, context information represents the global position of human and indicates the contextual relationship between keypoints. Spatial information  can provide detail information which is useful for refining the positions of keypoints. Thus,  the well-designed network should take both context information and spatial information into account. Several principles (e.g. preserving spatial information, and capturing diverse context information) proposed for designing networks in scene parsing can be also effective for the multi-person pose estimation task. 

\noindent{\textbf{Context Information}}~~Generally, as the network goes deep, the high-level feature holds potential to capture the context information with a large receptive filed. In another way, Atrous Spatial Pyramid Pooling (ASPP)~\cite{chen2018deeplab} and Pyramid Pooling Module (PPM)~\cite{zhao2017pyramid} are widely used to extract abundant context information in scene parsing.  ASPP module employs atrous convolution with different dilation rates  and global pooling module to capture diverse context information. PPM module fuses features under different pyramid pooling scales to obtain global contextual prior information. 

\noindent{\textbf{Spatial Information}}~~Consecutive down-sampling or pooling operations in the convolution neural network may lose the spatial information which is crucial to predicting the detailed output in scene parsing and pose estimation tasks.  Some existing methods~\cite{chen2017rethinking,zhao2017pyramid,chen2018deeplab,wang2018understanding} use the dilated convolution to preserve spatial size of the feature map. Other methods employ the feature pyramid network~\cite{lin2017feature}, U-shape method~\cite{ronneberger2015u}, Hourglass network~\cite{newell2016stacked} to shorten the information path between low-level features and high-level features. By using such skip-connected network structure, we can recover a certain extent of spatial information.

In our paper, we aim at leveraging both the context information and spatial information to improve multi-person pose estimation. Compared with existing methods, we design a Structure Supervision module to capture part-aware context information and adopt ASPP module to capture context information of different receptive fields in Context Aware Path. We preserve abundant spatial information by adding skip layers from the low-level features to high-level features with Spatial Aware Path. Experimentally, we find that adding the global pooling features of low-level features can further help accurately locate the keypoints. Moreover, we proposed a simple yet effective Heavy Head Path to fuse the context aware features and spatial aware features. 

\section{Method}
In this section, we propose a novel Context-and-Spatial Aware Network (CSANet) to make full use of the context information and spatial information. An overview of the proposed CSANet is illustrated in Figure~\ref{fig:image2}. We first briefly review the structure of Simple Baseline Network. Then, we  introduce Context-Aware-Path, Spatial-Aware-Path, and Heavy Head Path in detail. Finally, we describe the complete network architecture of Context-and-Spatial Aware Network, as well as training and inference details.

\subsection{Revisiting Simple Baseline Network}


ResNet is the most commonly used backbone network for image classification, scene parsing, and human pose estimation. Simple Baseline Network (SBN) uses a DeconvHead (consists of three deconvolution layers) after the last convolution stage of the ResNet, in which each deconvolution layer has 256 filters with $4\times4$ kernel size and stride parameter is 2. After the DeconvHead, a $1\times1$ convolution layer is added to predict the heatmaps for all keypoints. 

\begin{figure}[t]
\begin{center}
\includegraphics[height = 8cm]{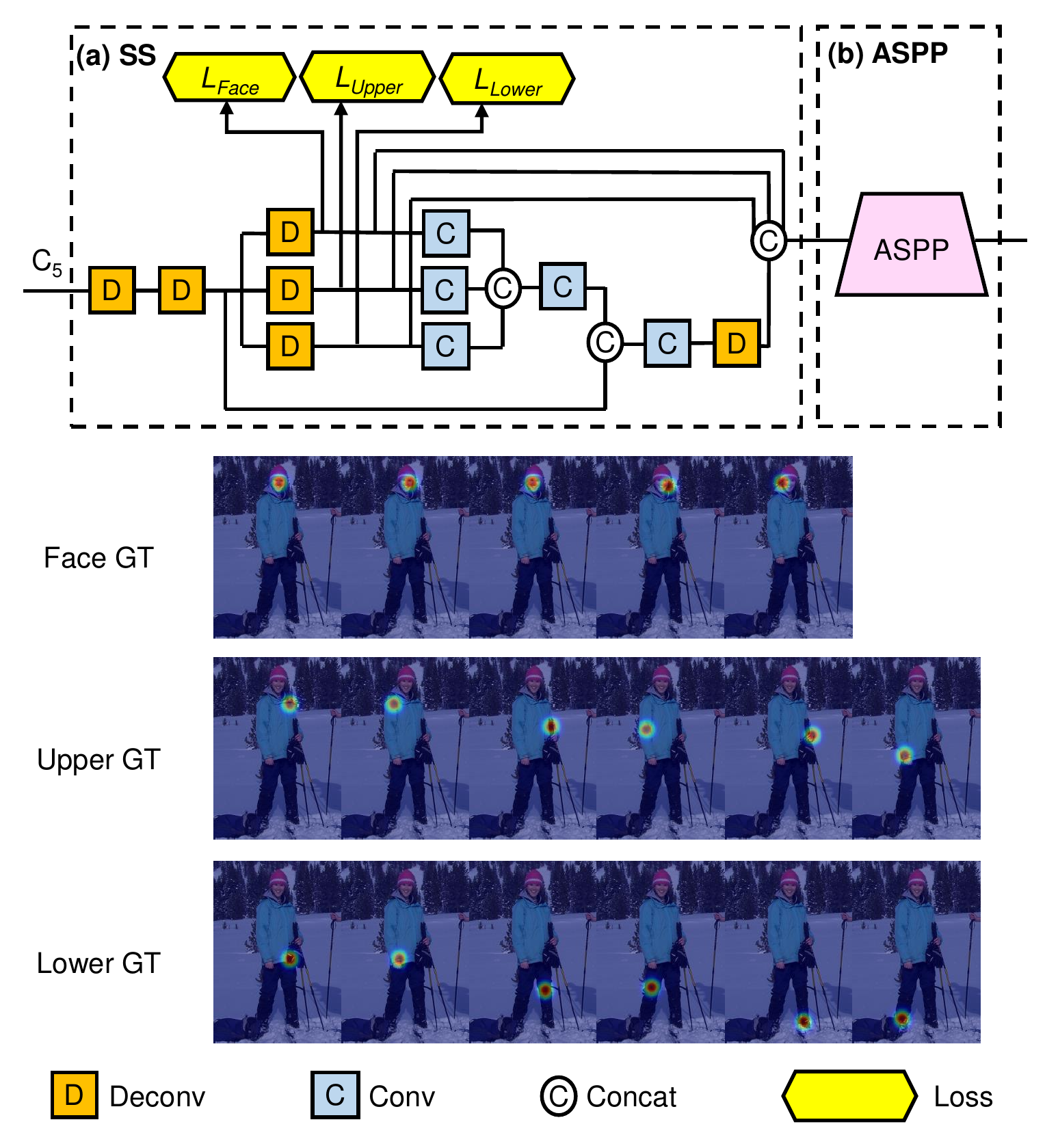}
\end{center}
   \caption{Illustration of the Context Aware Path. (a) is the Structure Supervision (SS) module. (b) is the Atrous Spatial Pyramid Pooling (ASPP) module. Face GT denotes the ground-truth score maps of the face related keypoints. Upper GT represents the ground-truth score maps of the upper limb related keypoints. Lower GT indicates the ground-truth score maps of the lower limb related keypoints. $L_{Face}$, $L_{Upper}$, and $L_{Lower}$ are the loss of the face, upper limb, and lower limb related keypoints, respectively.} 
\label{fig:image3}
\end{figure}

\subsection{Context Aware Path}

\noindent{\textbf{Motivation}}~~In the task of multi-person pose estimation, most of the modern methods tackle it as a dense regression issue. Due to the lack of abundant context information, the regression could not handle the prediction of invisible keypoints, occluded keypoints, and other complex situations. To this end, we design a Context Aware Path to extract abundant context information which represents the global position of human and  the contextual relationship of body keypoints.  

The Context Aware Path contains two modules: Structure Supervision module and Atrous Spatial Pyramid Pooling module, as shown in Figure~\ref{fig:image3}. The Structure Supervision (SS) module is to encode part-aware context information,  and the Atrous Spatial Pyramid Pooling (ASPP) module is to capture diverse context information of different receptive fields. 

\noindent{\textbf{Structure Supervision}}~~Body structural priors can provide valuable cues to infer the locations of the hidden body parts from the visible ones. Motivated by this, we perform multi-part supervision at each part prediction branch to obtain part-aware features. Compared with the Simple Baseline Network, we replace the DeconvHead module with the Structure Supervision module. In this paper, we divide human body into three parts for COCO keypoint dataset: face part (ears, eyes, and nose), upper limb part (shoulders, elbows, and wrists) and lower-limb part (hips, knees, and ankles). Then, we combine the face aware features, upper limb aware features, and lower limb aware features with the hybrid context features for capturing diverse part-aware context information. 

\noindent{\textbf{Atrous Spatial Pyramid Pooling}}~~ Atrous convolution is a powerful operation to adjust the filed-of-view in order to capture multi-scale information.  The Atrous Spatial Pyramid Pooling module has been widely used in scene parsing task, which adopts atrous convolution with different dilation rates for diverse context information extraction. In the CAP path, we simply add this module after the structure supervision module to capture context information of different receptive fields. In our experiment,  we set the dilation rates as 1, 6, 12, and 18.

As shown in Figure~\ref{fig:image3}, three branches are employed to extract part-aware context features, which respectively supervised by face ground-truth score maps, upper limb ground-truth score maps, and lower limb ground-truth score maps. Then ASPP module is adopted to further recalibrate the fusing information of  the part-aware features and hybrid context features.  

\subsection{Spatial Aware Path}\label{SAPMethod}

\noindent{\textbf{Motivation}}~~In the task of multi-person pose estimation, spatial information can provide detailed information which is useful for refining the positions of keypoints.  Some existing methods~\cite{cao2017realtime,insafutdinov2016deepercut,papandreou2017towards} attempt to estimate keypoints from the heatmaps of which the resolution is 1/8 of the input image. Yet, higher resolution information should be added to provide more spatial details.  To this end, we extract the spatial aware features from the lower stages of the backbone network to preserve abundant spatial information of which the resolution is 1/4 of the input resolution.

\begin{figure}[t]
\begin{center}
\includegraphics[height = 5cm]{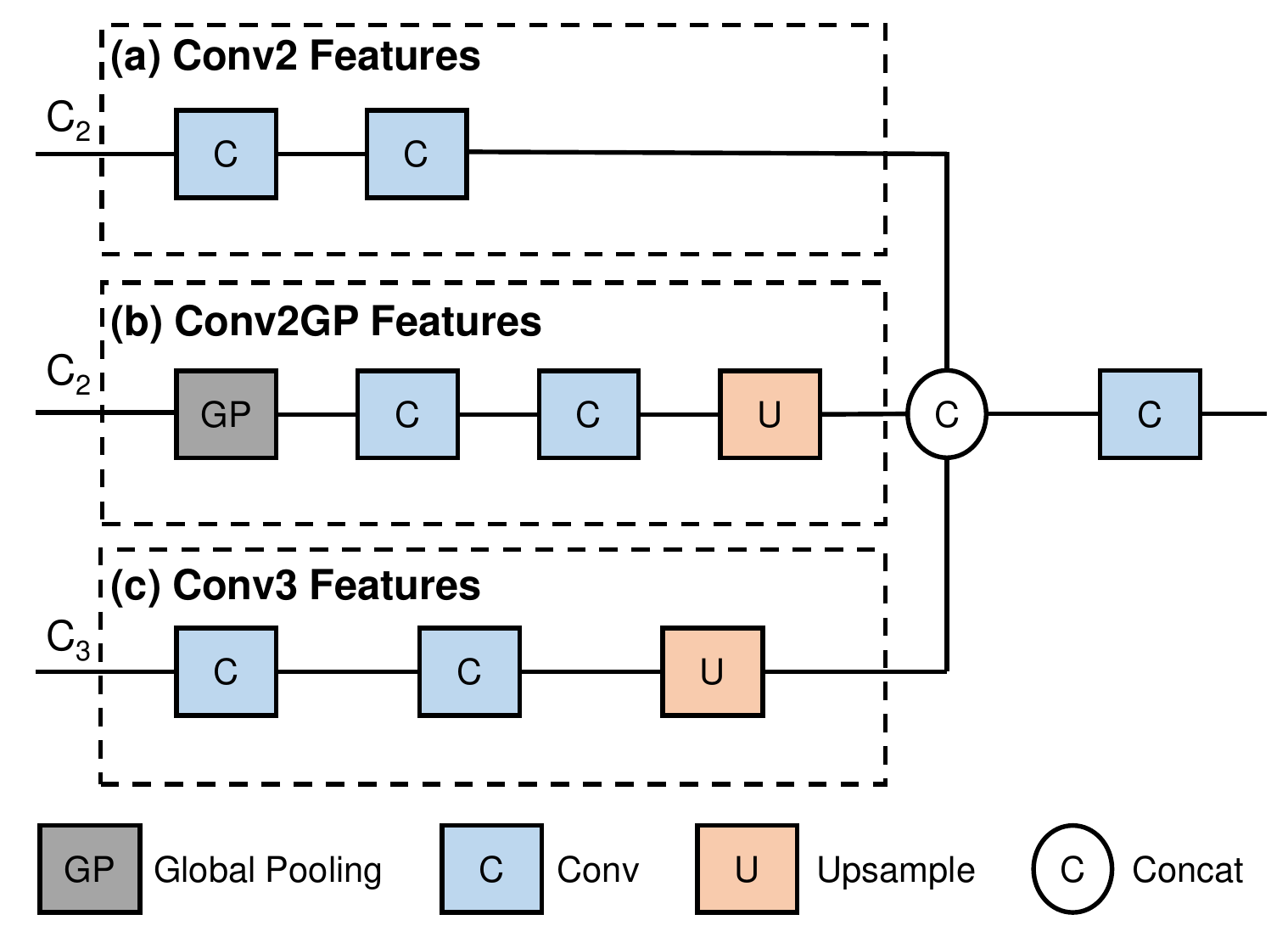}
\end{center}
   \caption{Illustration of the Spatial Aware Path. (a) is the branch to obtain Conv2 Features. (b) is the branch to obtain Conv2GP Features. (c) is the branch to obtain Conv3 Features.}
\label{fig:image4}
\end{figure}

In our proposed network, we use ResNet~\cite{he2016deep} as a backbone model. According to the feature maps' size, the ResNet can be divided into five stages, denoted as $\mathrm{C}_1$, $\mathrm{C}_2$, $\mathrm{C}_3$, $\mathrm{C}_4$, and $\mathrm{C}_5$ stages. The ResNet encodes more detailed spatial information in the lower stages, however, extracts stronger context information in the higher stages. Based on this observation, we design our Spatial Aware Path to capture the finer spatial information, as shown in Figure~\ref{fig:image4}. First, we use a convolution path (contains a convolution layer with $[256, 3\times3]$ filters and  a convolution layer with $[256, 1\times1]$ filters) to recalibrate the last feature maps of $\mathrm{C}_2$ stage to obtain the spatial feature, denoted as Conv2 Features. Conv3 Features are captured by another convolution path (same as Conv2 Features) on the last feature maps of $\mathrm{C}_3$ stage and resized to the resolution of Conv2 Features. Next, we use a series of operations (consists of global pooling, two convolution layers with $[256, 1\times1]$ filters, and resize to the resolution of Conv2 Features) to generate Conv2GP Features. Finally, we concatenate the Conv2 Features, Conv3 Features and Conv2GP Features, and reduce the concatenated features to 256 dimension feature maps by a convolution layer with $[256, 1\times1]$ filters. Experimentally, we find that adding the Conv2GP Features can further help accurately locate the keypoints. 

\subsection{Heavy Head Path}
Heavy head, namely stack of convolution layers,  is quite effective for bounding box prediction~\cite{liu2018path}. In our paper, we find it is also useful in the dense regression of keypoints' score maps.  This path first concatenates the context aware features extracted by Context Aware Path and spatial aware features captured by Spatial Aware Path, followed by using a  small fully convolution network (FCN) to regress the body keypoints' ground-truth score maps.  After concatenating the context aware features and spatial aware features, the fusion parameters of these features can be adaptively learned by  the network. The FCN consists of $N+1$ convolution layers. The first $N$ layers consists of $[256, 3\times3]$ filters and the last convolution layer is $[K,1\times1]$ filters. According to the number of keypoints in the COCO keypoint benchmark~\cite{cocodataset.org}, $K$ is set to 17 in our paper. 

\subsection{Network Architecture, Training and Inference}

With the Context Aware Path, Spatial Aware Path, and Heavy Head Path, we propose a novel Context-and-Spatial Aware Network (CSANet) for multi-person pose estimation as illustrated in Figure~\ref{fig:image2}.

\noindent{\textbf{Network Architecture}}~~We use the pre-trained ResNet as our backbone network. First, we operate the Context Aware Path (CAP) on the last feature maps of $\mathrm{C}_5$ stage to capture the part-aware context information and diverse context information of different receptive filed. Then, we employ the Spatial Aware Path (SAP) on the  last feature maps of $\mathrm{C}_2$, $\mathrm{C}_3$ stage to encode spatial information feature. In our paper, the CAP path encodes abundant context information, while the SAP path provides rich spatial information. They are complementary to each other for higher performance on keypoint localization. Given the different feature representation of the CAP path and SAP path, we concatenate these features instead of simply summing operation to fuse the context information features and spatial information features. Finally, the Heavy Head Path (HHP) is operated on the concatenated features, which encodes both affluent context information and spatial information to accurately predict the keypoints' heatmaps.

\noindent{\textbf{Network Training}}~~In our paper, we use the score maps to represent the location of body keypoints. For each person, the ground-truth locations are labeled as $\bm{z} = \left\{ \bm{z}_k \right\}_{k=1}^K$, where $\bm{z}_k = (x_k, y_k)$ denotes the coordinate of the $k$th keypoint ($k =1,\cdots,5$ denote the keypoints of the face, $k= 6,\cdots,11$ denote the keypoints of the upper limb, and $k = 12,\cdots,17$ denote the keypoints of the lower limb) of the person. The ground-truth score map $\mathbf{G}_k$ is defined as, 
\begin{equation}
\mathbf{G}_k(\bm{r}) = exp(-\frac{||\bm{r}-\bm{z}_k||^2}{2\sigma^2}) ,
\end{equation}
in which, $\bm{r} \in R^2$ denotes the location, and $\sigma$ is set to 2 for $256\times192$ input, and is set to 3 for $384\times288$ input. In our Context Aware Path, we use three auxiliary loss functions to supervise it to learn part-aware context information. The face aware branch, upper limb aware branch and lower limb branch predict the face related keypoints' heatmaps ( \textit{i.e.}, $\mathbf{O}=\left\{ \mathbf{O}_k \right\}_{k=1}^5$), upper limb related keypoints' heatmaps ( \textit{i.e.}, $\mathbf{P}=\left\{ \mathbf{P}_k \right\}_{k=1}^6$), and lower limb related keypoints' heatmaps ( \textit{i.e.}, $\mathbf{Q}=\left\{ \mathbf{Q}_k \right\}_{k=1}^6$) respectively. For the Heavy Head Path, it predicts the holistic body keypoints' heatmaps ( \textit{i.e.}, $\mathbf{S}=\left\{ \mathbf{S}_k \right\}_{k=1}^{17}$). Then, the loss of our CSANet is, 
\begin{gather}
L_{face} = \frac{1}{2}\sum\limits_{i=1}^{M}\sum\limits_{k=1}^{5}||\mathbf{G}_k-\mathbf{O}_k||^2 \\
L_{upper} = \frac{1}{2}\sum\limits_{i=1}^{M}\sum\limits_{k=1}^{6}||\mathbf{G}_{k+5}-\mathbf{P}_k||^2 \\
L_{lower} = \frac{1}{2}\sum\limits_{i=1}^{M}\sum\limits_{k=1}^{6}||\mathbf{G}_{k+11}-\mathbf{Q}_k||^2 \\
L_{body} = \frac{1}{2}\sum\limits_{i=1}^{M}\sum\limits_{k=1}^{17}||\mathbf{G}_k-\mathbf{S}_k||^2 \\
L_{all} = \alpha L_{face} +\beta L_{upper} + \gamma L_{lower} + L_{body}
\end{gather}
where $M$ is the number of samples, $\alpha$, $\beta$, and $\gamma$ are the loss weight parameters.

\noindent{\textbf{Network Inference}}~~During the inference, we obtain the predicted body keypoints localizations $\hat{\bm{z}_k}$ from the predicted score maps generated from the Heavy Head Path by taking the locations with the maximum score as follows:
\begin{equation}
\hat{\bm{z}_k} = \mathop{\arg\max}_{\bm{r}} \hat{\mathbf{S}}_k(\bm{r}), k = 1,\cdots,K.
\end{equation}

\section{Experiment}

This section is organized in accordance with the progress of our experiments. Firstly, we describe the experimental setup. Then, we decompose our proposed network to reveal the effect of each component on MS-COCO val2017 dataset. Last but not least, we compare our network with previous state-of-the-art methods on MS-COCO val2017 dataset and MS-COCO test-dev2017 dataset.

\subsection{Experimental Setup}
\noindent{\textbf{Dataset and Evaluation Metric}}~~We train and evaluate our Context-and-Spatial Aware Network (CSANet) on MS-COCO 2017 dataset~\cite{lin2014microsoft}.  Our models are only trained on the MS-COCO train2017 dataset including 57K images and 150K person instances, no extra data involved. There are 5000 images (MS-COCO val2017 dataset) for validation and 20K images (MS-COCO test-dev2017 dataset) for testing. Following previous work~\cite{cao2017realtime,chen2017cascaded,xiao2018simple}, evaluation is conducted using the Object Keypoints Similarity (OKS) based mAP, where OKS defines the  difference between predicted person keypoints and ground-truth person keypoints.

\noindent{\textbf{Cropping Strategy}}~~The person ground-truth box (or detection box) is changed to a fixed aspect ratio, \textit{e.g.} height : weight = 4 : 3.  Then, we crop the image and resize it to a fixed resolution. In our paper, the default resolution of the network input image is $256\times192$. 

\noindent{\textbf{Data Augmentation Strategy}}~~We use random flip, random scale, and random rotation in training. The possibility of flip or not is 0.5. The random rotation range is ($-40^o \sim 40^o$), and the random scale is ($0.7 \sim 1.3$).   

\noindent{\textbf{Person Detector}}~~For MS-COCO val2017 dataset, we use the human detection boxes provided by~\cite{xiao2018simple} to make a fair comparison, the detection boxes are generated by a Faster-RCNN detector~\cite{ren2015faster} with human detection AP 56.4 on MS-COCO val2017. For MS-COCO test-dev2017 dataset, we adopt the SNIPER detector~\cite{singh2018sniper} with human detection AP 58.1 on MS-COCO test-dev2017.

\begin{table*}[tb]
    \renewcommand\arraystretch{0.9}
	\centering
	\caption{Results on the MS-COCO val2017 dataset. Based on the ResNet-50, we gradually add Context Aware Path (CAP), Spatial Aware Path (SAP), and Heavy Head Path (HHP) for ablation study. The first row is performance of Simple Baseline Network (SBN) which is the state-of-the-art performance network on COCO keypoint benchmark.}
    \label{table:table1}
	\begin{tabular}{cccccc}
		\hline
		Method & AP & AP.5 & AP.75 & AP(M)& AP(L) \\
		\hline
		ResNet-50+DeconvHead (SBN) & 70.6 & - & - & - & -\\
		ResNet-50+CAP & 71.1 & 88.8 & 78.6 & 67.7 & 77.6\\
		ResNet-50+CAP+SAP & 71.7 & 88.8 & 78.8 & 68.2 & 78.4 \\
		\textbf{ResNet-50+CAP+SAP+HHP (Our CSANet) }& \textbf{72.5} & \textbf{89.4} & \textbf{79.4} & \textbf{69.1} & \textbf{79.4}\\
		\hline
	\end{tabular}
\end{table*}

\begin{table}[tb]
    \renewcommand\arraystretch{0.9}
	\centering
	\caption{Ablation study on our proposed Context Aware Path. \textbf{SBN}: Simple Baseline Network. \textbf{SS:} Structure Supervision module. \textbf{ASPP:} Atrous Spatial Pyramid Pooling. }
    \label{table:table2}
	\begin{tabular}{cc}
		\hline
		Method & AP  \\
		\hline
		ResNet-50+DeconvHead (SBN) & 70.6 \\
		ResNet-50+SS & 71.0  \\
		\textbf{ResNet-50+SS+ASPP} & \textbf{71.1} \\
		\hline
	\end{tabular}
\end{table}

\noindent{\textbf{Training Details}}~~We train our proposed CSANet using Adam~\cite{kingma2014adam} algorithm with a mini-batch of 128 (32 per GPU) for 140 epochs. The initial learning rate is 1e-3 and is dropped by 10 at the 90th epoch and the 120th epoch. Generally, the training of ResNet-50 based models takes about 52 hours on four NVIDIA Titan V100 GPUs. All codes are implemented with PyTorch~\cite{paszke2017automatic}. In this paper, our network is trained with ResNet-50, ResNet-101, and ResNet-152. The ResNet backbones are initialized with the public-released pre-trained model on the ImageNet~\cite{russakovsky2015imagenet}. We also conduct experiments with different resolutions of the input image ($256\times192$ and $388\times284$).    

\noindent{\textbf{Testing Details}}~~A top-down pipeline is adopted for estimating the multi-person pose. First, we use a person detector to generate the human bounding boxes. Then, we apply our CSANet to generate the pose prediction heatmaps for each bounding box. Following previous work~\cite{xiao2018simple,chen2017cascaded},  we average the heatmaps of origin image and the heatmaps of the flipped image to get the final prediction. A quarter offset in the direction from the highest response to the second highest response is used to obtain the final location.

\subsection{Ablation Study}
In this subsection, we will step-wise decompose our proposed CSANet to reveal the effect of each component. In the following experiments, we evaluate all comparisons on MS-COCO val2017 dataset.  Unless otherwise specified, the default backbone is ResNet-50, and the input size of all models is $256 \times 192$.

\subsubsection{Component Analysis}
In Table~\ref{table:table1}, we show our ablation study from the Simple Baseline Network~\cite{xiao2018simple} (SBN, which achieves the state-of-art) gradually to all components incorporated. Based on the SBN, we replace the DeconvHead with our Context Aware Path (CAP), the AP performance is improved from 70.6 to 71.1. Furthermore, when adding the Spatial Aware Path, we can achieve 71.7 AP. Finally, we adopt the Heavy Head Path (HHP) to fuse the context  aware information and spatial aware information to predict the pose heatmaps. After adding the HHP module, the AP performance can be further improved from 71.7 to 72.5.  

\subsubsection{Ablation Study on Context Aware Path}
Different with the SBN, we replace the DeconvHead (consists of three deconvolution layers) with our Context Aware Path. The CAP path consists of two modules: Structure Supervision module and ASPP module. 

\noindent{\textbf{Ablation for Structure Supervision}}~~We use the Structure Supervision module which performs multi-part supervision operation to extract the part-aware context information. As shown in Table~\ref{table:table2}, this module improves AP performance from 70.6 to 71.0, which is an obvious improvement. In our paper, the loss weight parameters $\alpha$, $\beta$, and $\gamma$ are set to 1. We also conduct the experiment which sets all the loss weight parameters to 0, the AP performance is 70.8. 

\noindent{\textbf{Ablation for Atrous Spatial Pyramid Pooling}}~~To capture diverse context information of different receptive fields, we apply the ASPP module on the features extracted by Structure Supervision module. As shown in Table~\ref{table:table2}, this further improves the performance by 0.1.

\begin{table}[tb]
    \renewcommand\arraystretch{0.9}
	\centering
	\caption{Ablation study on our proposed Spatial Aware Path. \textbf{Conv2:} Features captured from $\mathrm{C}_2$ stage. \textbf{Conv3:} Features captured from $\mathrm{C}_3$ stage. \textbf{Conv2GP:} Features captured from global pooling of $\mathrm{C}_2$ stage.}
    \label{table:table3}
	\begin{tabular}{cc}
		\hline
		Method & AP  \\
		\hline
		ResNet-50+CAP & 71.1 \\
		ResNet-50+CAP+Conv2 & 71.4  \\
		ResNet-50+CAP+Conv2+Conv3 & 71.5 \\
        ResNet-50+CAP+Conv2+Conv2GP & 71.4 \\
        \textbf{ResNet-50+CAP+Conv2+Conv3+Conv2GP} & \textbf{71.7} \\ 
		\hline
	\end{tabular}
\end{table}

\subsubsection{Ablation Study on Spatial Aware Path}

While the Context Aware Path pays attention to the context information, the Spatial Aware Path focus on the spatial information which can provide detail information for refining the positions of keypoints. By integrating the CAP path and SAP path, the AP performance is improved from 71.1 to 71.7, as shown in Table~\ref{table:table3}.  

\noindent{\textbf{Design Choices of Spatial Aware Path}}~~Here, we compare different design strategies of the SAP path as shown in Table~\ref{table:table3}. We compare the following implementations: 1) Conv2 Features. 2) Conv2 Features + Conv3 Features. 3) Conv2 Features + Conv2GP Features. 4) Conv2 Features + Conv3 Features + Conv2GP Features. The Conv2 Features, Conv2GP Features, and Conv3 Features are  detailedly described in Section~\ref{SAPMethod}. 


Then, we reduce the spatial aware features to 256 dimension, and integrating it with the context aware features to predict the person pose. As shown in Table~\ref{table:table3}, we find that adding spatial detail information with Conv2 Features, Conv3 Features and Conv2GP Features can effectively achieve 0.6 AP gains.

\begin{figure*}
\begin{center}
\includegraphics[width = 15.5cm, height = 7.45cm]{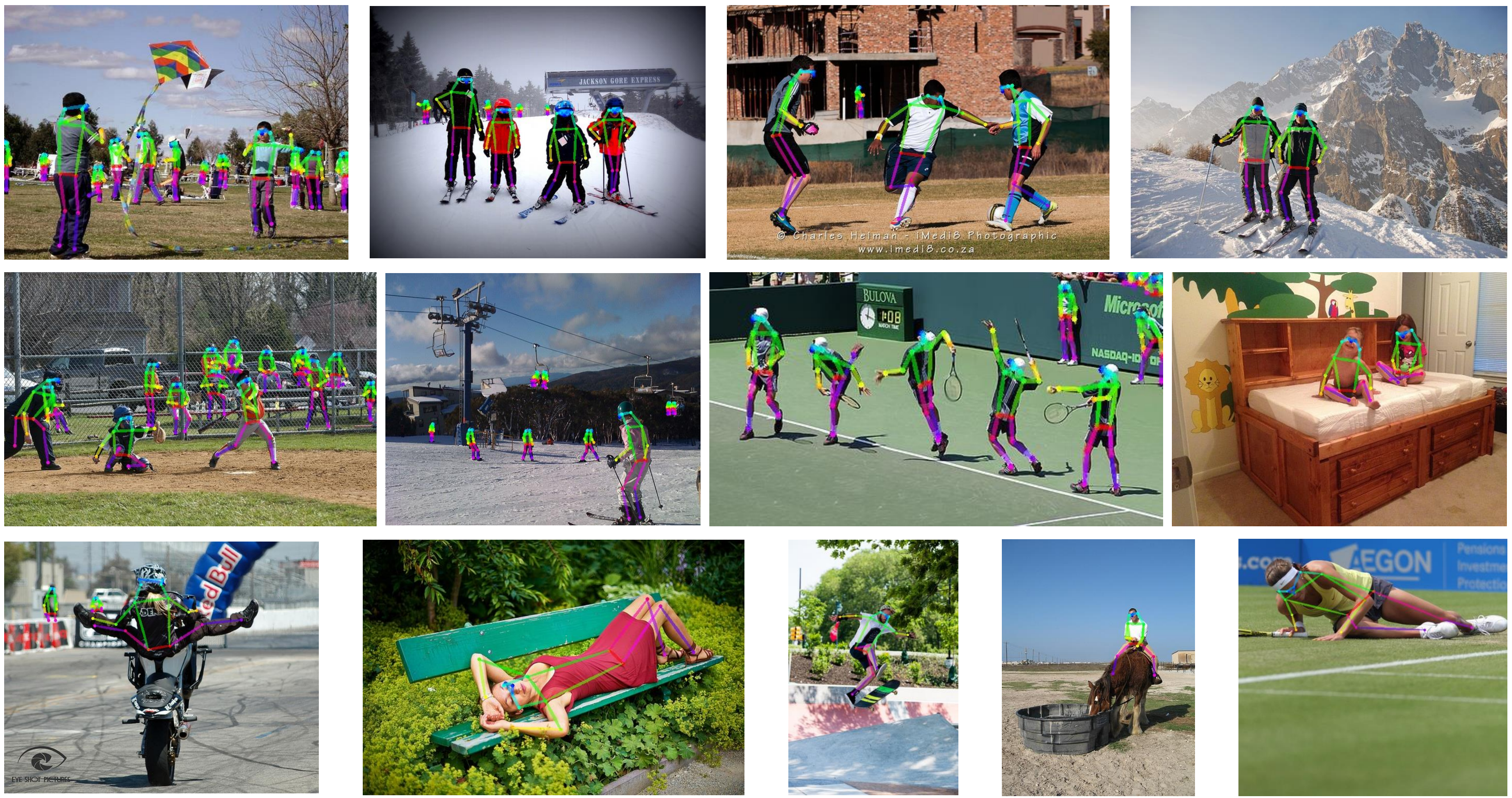}
\end{center}
   \caption{Some results from MS-COCO test-dev 2017 dataset of our method.}
\label{fig:image5}
\end{figure*}

\subsubsection{Ablation Study on Heavy Head Path}

This path first concatenates the context aware features extracted by Context Aware Path and spatial aware features captured by Spatial Aware Path, then uses a  small fully convolution network (FCN) to regress the keypoints' ground-truth score maps. The FCN consists of N convolution layers with $[256, 3\times3]$ filters and one convolution layer with $[17, 1\times1]$ filters. As shown in Table~\ref{table:table4},  this improves the AP performance from 71.7 to 72.5 when N is chosen as 3, 5, or 6. In our CSANet, the N is set to 3 for less computation. 

\begin{table}[tb]
    \renewcommand\arraystretch{0.9}
	\centering
	\caption{Ablation study on our proposed Heavy Head Path. N denotes the number of convolution layers used in this module. The baseline is ResNet-50+CAP+SAP model.}
    \label{table:table4}
	\begin{tabular}{cccccccc}
		\hline
		N & 0  & 1 & 2 & \textbf{3 }& 4 & 5 & 6 \\ 
 		\hline
		AP & 71.7 & 71.9 & 72.0 & \textbf{72.5} & 72.4 & 72.5 & 72.5 \\
		\hline
	\end{tabular}
\end{table}

\begin{table}[tb]
    \renewcommand\arraystretch{0.9}
	\centering
	\caption{Ablation study of different input sizes.}
    \label{table:table5}
	\begin{tabular}{cccc}
		\hline
		Method & Input Size & Backbone & AP  \\
		\hline
        Our CSANet  & $256 \times 192$ & ResNet-50 & 72.5 \\
        \textbf{Our CSANe}t & $384 \times 288$ & ResNet-50 & \textbf{74.1} \\
		\hline
	\end{tabular}
\end{table}

\begin{table}[tb]
    \renewcommand\arraystretch{0.9}
	\centering
	\caption{Ablation study of different Backbone networks.}
    \label{table:table6}
	\begin{tabular}{cccc}
		\hline
		Method & Backbone & Input Size & AP  \\
		\hline
        Our CSANet  &  ResNet-50 & $384 \times 288$ & 74.1 \\
        Our CSANet  &  ResNet-101 & $384 \times 288$ & 74.4 \\
        \textbf{Our CSANet}  &  ResNet-152 & $384 \times 288$ & \textbf{75.1 }\\
		\hline
	\end{tabular}
\end{table}

\begin{table}[tb]
    \renewcommand\arraystretch{0.9}
	\centering
	\caption{Comparison with Hourglass~\cite{newell2016stacked}, CPN~\cite{chen2017cascaded}, SBN~\cite{xiao2018simple} on MS-COCO val2017 dataset. \textbf{Hourglass:} a classical model. \textbf{CPN:} Cascade Pyramid Network, COCO2017 keypoint winner. \textbf{SBN:} Simple Baseline Network, the state-of-the-art network. }
    \label{table:table7}
	\begin{tabular}{cccc}
		\hline
		Method & Backbone & Input Size & AP \\
		\hline
        8-stage Hourglass &  ResNet-50 & $256 \times 192$ & 66.9 \\
        8-stage Hourglass &  ResNet-50 & $256 \times 256$ & 67.1 \\
        CPN  &  ResNet-50& $256 \times 192$ & 69.4 \\
        CPN  &  ResNet-50 & $384 \times 288$ & 71.6\\
        SBN  &  ResNet-50 & $256 \times 192$ & 70.6 \\
        SBN  &  ResNet-50 & $384 \times 288$ & 72.2\\
		\hline
        \textbf{Our CSANet}  &  ResNet-50 & $256 \times 192$ & \textbf{72.5} \\
        \textbf{Our CSANet } &  ResNet-50 & $384 \times 288$ & \textbf{74.1} \\
        \hline
	\end{tabular}
\end{table}

\subsubsection{Ablation Study on Data Pre-processing}
Here, we investigate the performance of our CSANet with different input sizes. Due to the increase of input image size, more spatial information are fed into our network. Therefore, this improves the AP performance from 72.5  ($256\times192$ input size) to 74.1 ($388\times284$ input size), which is an obvious large improvement, as shown in Table~\ref{table:table5}.

\subsubsection{Ablation Study on Backbone Network}
As in most computer vision tasks, a deeper backbone model has better performance.  We conduct experiments with ResNet-50, ResNet-101, and ResNet-152 backbones with the input size of $384\times288$.  Table~\ref{table:table6} shows that AP increase is 0.3 from ResNet-50 to ResNet-101 and 1.0 from ResNet-50 to ResNet-152.

\begin{table*}[tb]
   \renewcommand\arraystretch{0.9}
	\centering
	\caption{Comparisons on the MS-COCO test-dev2017 dataset. \textbf{Top:} methods in the literature, trained only on COCO training dataset. \textbf{CMU-Pose:} COCO2016 keypoint winner~\cite{cao2017realtime}. \textbf{Mask-RCNN:} a classical model~\cite{he2017mask}. \textbf{G-RMI:} a classical model~\cite{papandreou2017towards}. \textbf{CPN:} Cascaded Pyramid Network, COCO2017 keypoint winner~\cite{chen2017cascaded}. \textbf{SBN: }Simple Baseline Network, the state-of-the-art network~\cite{xiao2018simple}. "+" means the method using ensemble models. \textbf{Bottom:} our single model results, trained only on COCO training dataset.}
    \label{table:table8}
	\begin{tabular}{ccccccccc}
		\hline
		Method & Backbone & Input Size & AP & AP.5 & AP.75 & AP(M)& AP(L) & AR \\
		\hline
        CMU-Pose & - & -  &61.8&84.9&67.5&57.1&68.2&66.5\\
        Mask-RCNN & ResNet-50-FPN & - &63.1&87.3&68.7&57.8&71.4&-\\
        G-RMI & ResNet-101 & $353\times257$ &64.9&85.5&71.3&62.3&70.0&69.7\\
        CPN & ResNet-Inception & $384\times288$ &72.1&91.4&80.0&68.7&77.2&78.5\\
        CPN+ & ResNet-Inception & $384\times288$ &73.0&91.7&80.9&69.5&78.1&79.0\\
        SBN & ResNet-50& $256\times192$ &70.2&90.9&78.3&67.1&75.9&75.8\\
        SBN & ResNet-50& $384\times288$ &71.3&91.0&78.5&67.3&77.9&76.6\\
        SBN & ResNet-101& $256\times192$ &71.1&91.1&79.3&68.3&76.7&76.8\\
        SBN & ResNet-101& $384\times288$ &73.2&91.4&80.9&69.7&79.5&78.6\\
        SBN & ResNet-152& $256\times192$ &71.9&91.4&80.1&68.9&77.4&77.5\\
        SBN & ResNet-152& $384\times288$ &73.8&91.7&81.2&70.3&80.0&79.1\\
        \hline
        Our CSANet & ResNet-50 &  $256\times192$ &71.9&91.0&79.9&68.7&77.5&78.7 \\
        Our CSANet& ResNet-50 &  $384\times288$ &73.5&91.4&80.8&69.9&79.4&79.7\\
        Our CSANet & ResNet-101 &  $256\times192$ &72.3&91.2&80.2&69.3&77.6&79.1\\
        Our CSANet & ResNet-101 &  $384\times288$ &74.1&91.6&81.6&70.7&79.8&80.4\\
        Our CSANet & ResNet-152 &  $256\times192$ &72.8&91.4&80.9&69.8&78.3&79.6\\
        \textbf{Our CSANet} & ResNet-152 &  $384\times288$ &\textbf{74.5}&\textbf{91.7}&\textbf{82.1}&\textbf{71.2}&\textbf{80.2}&\textbf{80.7}\\
		\hline
	\end{tabular}
\end{table*}

\subsection{Comparison with State-of-the-art Methods}
In this subsection, we compare our proposed CSANet with state-of-the-art methods on MS-COCO val2017 dataset and MS-COCO test-dev2017 dataset. 

\noindent{\textbf{Results on MS-COCO val2017}} As shown in Table~\ref{table:table7}, we compare our network with a 8-stage Hourglass (a classical model), CPN (Cascaded Pyramid Network, COCO2017 winner), and SBN (Simple Baseline Network, the state-of-the-art network) . All these methods use top-down pipeline. For  human bounding boxes generating, the person detection AP of Hourglass and CPN is 55.3. The person detection AP of SBN is 56.4, we use the human bounding boxes provided by SBN to make a fair comparison.  

Compared with Hourglass~\cite{newell2016stacked}, our CSANet has an improvement of 5.6 points in AP for input size of $256 \times 192$. Our network outperforms CPN~\cite{chen2017cascaded} by 3.1 AP for input size of $256 \times 192$, and 2.5 AP for input size of $384 \times 288$. By contrasting SBN~\cite{xiao2018simple} with our CSANet, the AP performance is improved from 70.6 to 72.5 for input size of $256 \times 192$, and from 72.2 to 74.1 for input size of $384 \times 288$.  Our method improves the previous best results with a large margin by 1.9 AP for both $256 \times 192$ and $384 \times 288$ input size.

\noindent{\textbf{Results on MS-COCO test-dev 2017}} Table~\ref{table:table8} illustrates the results of modern state-of-the-art methods in the literature on MS COCO test-dev2017 dataset. For the human bounding boxes generating, CPN uses a human detector with  person detection AP 62.9 on COCO minival split dataset. SBN adopts a human detector with person detection AP 60.9 on COCO test-dev dataset. We use the SNIPER detector with person detection AP 58.1 on COCO test-dev dataset.

Compared with CMU-Pose~\cite{cao2017realtime}, G-RMI~\cite{papandreou2017towards}, and Mask-RCNN~\cite{he2017mask}, our method achieves significant improvement. Even though CPN~\cite{chen2017cascaded} use a stronger backbone of ResNet-Inception, our CSANet's single model (ResNet-152) achieves 74.5 AP and outperforms CPN's single model by 2.4 AP for the input size of $384 \times 288$. As mentioned before,  SBN~\cite{xiao2018simple} use  a more powerful human detector with person detection AP 60.9 on COCO test-dev dataset, which is higher than our human detector by 2.8. Yet, our model has an improvement of 0.7 AP in multi-person pose estimation for the input size of $384 \times 288$. Figure~\ref{fig:image5} illustrates some results generated using our method.

\section{Conclusion}
Aiming at fully leveraging both context information and spatial information to improve multi-person pose estimation, we propose a novel Context-and-Spatial Aware Network (CSANet) in this paper. From the architecture perspective, we design a Context Aware Path to capture part-aware information and diverse context information of different receptive filed which  indicates the contextual relationship between keypoints. Then,  we propose a Spatial Aware Path to preserve detail information for refining the position of keypoints. Next, a Heavy Head Path is proposed to further combine and recalibrate the context aware features and spatial aware features. These modules are trained as a whole to maximally complement each other. We also conduct a series of ablation studies to validate the effectiveness of each module. Finally, our experimental results show that our proposed CSANet can significant improve the performance on COCO keypoint benchmark.

{\small
\bibliographystyle{ieee}
\bibliography{egbib}
}

\end{document}